\documentclass{article}
\usepackage{amsmath,graphicx,mlspconf}
\usepackage{xcolor}
\usepackage{hyperref}
\usepackage{amsmath,amssymb,amsfonts}
\usepackage{algorithm}
\usepackage{algorithmicx}
\usepackage{algpseudocode}  

\usepackage{textcomp}
\usepackage{xcolor}
\usepackage{lscape,lipsum}
\usepackage{threeparttable}
\usepackage{longtable}
\usepackage{multirow}
\usepackage{makecell}
\usepackage{etoolbox}
\usepackage{pgfplots} 
\usepackage{tikz}
\usepackage{graphicx, adjustbox}
\usepackage{subfig}
\usepackage{mathtools}
\usepackage{array}
\usepackage{booktabs}
\usepackage[usestackEOL]{stackengine}
\usepackage{xfrac}
\usepackage{threeparttable}
\usetikzlibrary{shapes.geometric}
\usetikzlibrary{3d}
\usepackage{url}
\usepackage{mathtools}
\title{Shapley-Based Data Valuation with Mutual Information:\\ A Key to Modified K-Nearest Neighbors}
\name{Mohammad Ali Vahedifar$^{1}$, Azim Akhtarshenas$^{2}$, Mohammad M.RafatPanah$^{3}$, and Maryam Sabbaghian$^{3}$\thanks{Authors' e-mails: av@ece.au.dk, aakhtar@doctor.upv.es, mohammad.rafat@ut.ac.ir, msabbaghian@ut.ac.ir.\\Link to the code: \href{https://github.com/Ali-Vahedifar/Shapley-Value-Based-KNN-IMKNN.git}{GitHub Repository}.}}
\address{$^{1}$Department of Electrical and Computer Engineering, Aarhus University, Denmark\\
$^{2}$Department of Electrical and Computer Engineering, Universitat Polit\`ecnica de Val\`encia, Spain\\
$^{3}$Department of Electrical and Computer Engineering, University of Tehran, Iran}

\begin{document}
\maketitle
\begin{abstract}
The K-Nearest Neighbors (KNN) algorithm is widely used for classification and regression; however, it suffers from limitations, including the equal treatment of all samples. We propose Information-Modified KNN (IM-KNN), a novel approach that leverages Mutual Information ($\mathcal{I}$) and Shapley values to assign weighted values to neighbors, thereby bridging the gap in treating all samples with the same value and weight. On average, IM-KNN improves the accuracy, precision, and recall of traditional KNN by 16.80\%, 17.08\%, and 16.98\%, respectively, across 12 benchmark datasets. Experiments on four large-scale datasets further highlight IM-KNN's robustness to noise, imbalanced data, and skewed distributions.

\end{abstract}

\newcommand{\cem}[1]{\textcolor{blue}{cem: #1}}
\section{Introduction}
\label{sec:intro}
Pattern recognition is a fundamental area within Machine Learning (ML) that focuses on automatically identifying and classifying patterns in data~\cite{9838718}. The objective is to develop algorithmic frameworks and computational models that extract meaningful patterns, enabling systems to analyze and generalize from complex datasets. This process involves assigning labels to objects based on measurable features, which serve as distinguishing characteristics~\cite{Darasay}.

Among various ML techniques, the K-Nearest Neighbors (KNN) algorithm is widely employed for classification and regression due to its simplicity and effectiveness. As an instance-based learning method, KNN makes predictions by evaluating the similarity between a query point and labeled instances in the training set. The algorithm assigns labels to new data points based on the majority vote among the nearest neighbors. It operates without prior knowledge of the underlying data distribution and remains non-parametric~\cite{BANSAL2022100071}. Despite its advantages, traditional KNN treats all data points equally without considering the individual value of each data point~\cite{6618910}. 

\begin{figure}[t]
    \centering
    \includegraphics[width=\linewidth]{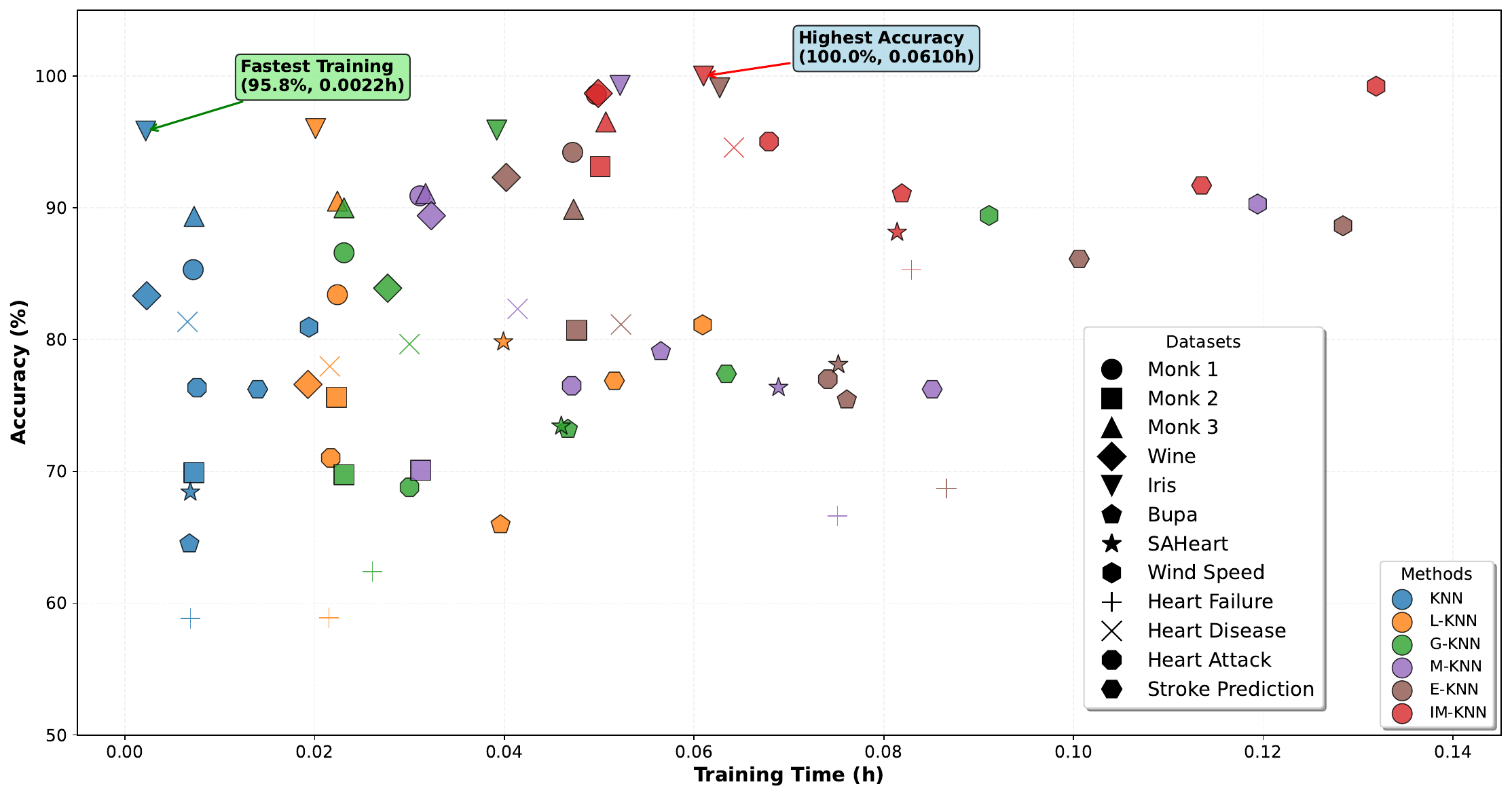}
    \caption{Training Time vs Accuracy of all methods for small-scale datasets.}
    \label{fig:timevsaccuracy}
\end{figure}

\begin{figure*}[htbp]
    \centering
    \includegraphics[width=\textwidth]{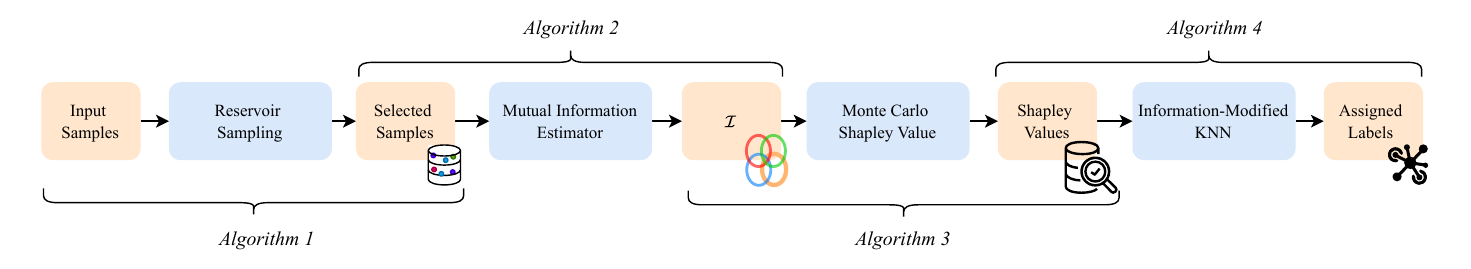}
    \caption{The process of IM-KNN flows sequentially: an initial reservoir sampling step reduces the input data. Next, a mutual information estimator computes $\mathcal{I}$ from this sampled subset. These $\mathcal{I}$ values then inform the calculation of Shapley values via a Monte Carlo approach. Ultimately, the derived Shapley values are applied within the IM-KNN algorithm to refine the weighting of neighbors. The output of each preceding stage serves as the direct input for the following stage.}
    \label{fig:IMKNN}
\end{figure*}

To mitigate this challenge, we develop an Information-Modified KNN (IM-KNN) method by incorporating data valuation into the majority voting process. By using Mutual Information ($\mathcal{I}$)-based performance metrics, we systematically compute Shapley values to quantify the contributions of data points~\cite{shapley1951notes}. Additionally, we introduce a distance-aware weighting mechanism for majority voting, where votes are weighted by the proximity of the test sample to its neighbors. This improves classification accuracy by emphasizing more informative data points. The result in Fig.~\ref{fig:timevsaccuracy} shows IM-KNN outperforms other methods in different datasets with comparable training time. The multi-stage conceptual pipeline of the IM-KNN method is presented in Fig~\ref{fig:IMKNN}.
\section{Related Works}
The KNN algorithm has been extensively studied and extended to address its limitations. Locally adaptive KNN (L-KNN) optimizes $K$ by analyzing the distribution of majority and second-majority class neighbors, ranking $K$ values based on centroid distances and ratios~\cite{pan2020}. Generalized mean distance KNN (G-KNN) employs local mean vectors and generalized mean distances to classify test points~\cite{gou2019}. Mutual KNN (M-KNN) filters training data to retain mutual nearest neighbors and classifies using majority voting~\cite{dhar2020}. Ensemble KNN (E-KNN) combines multiple KNN classifiers with varying $K$ values, weighting neighbors inversely by distance~\cite{hassanat2014solving}. In addition, Shapley value, a concept from cooperative game theory, has recently gained attention in ML for quantifying feature importance in black-box models~\cite{ghorbani19c, Lundberg2018, Hamers2016}. Our work diverges from these approaches by leveraging the Shapley value to evaluate the contributions of data points in a KNN context.

\section{INFORMATION MODIFIED K-Nearest Neighbors}

\textbf{Preliminaries:} Let \( D = \{(x_i, y_i)\}_{i=1}^{N} \) be a fixed training dataset, where \( x_i \) represents the input features and \( y_i \) denotes the corresponding labels. We make no explicit assumptions regarding the underlying distribution of \( D \), and the data points are not necessarily independent.
Let \( \mathcal{K} \) be a KNN learning algorithm that takes a training dataset of arbitrary size (\( 0 \leq |K| \leq N \)) as input and returns a trained predictor. Specifically, we want to analyze the predictor trained on subsets \( K \subseteq D \). To evaluate the quality of a predictor, we introduce a performance measure \( \mathcal{I} \), which is the mutual information oracle. This function, denoted as \( \mathcal{I}(K, \mathcal{K}) \), assigns a performance score to the predictor trained on subset \( K \). The objective is to compute a data valuation function \( \mathcal{X}_i(D, \mathcal{K}, \mathcal{I}) \in \mathbb{R} \) that quantifies the contribution of the \( i \)-th data point within the dataset. We may write this for notational simplicity as \(\mathcal{X}_i(\mathcal{I})\). Additionally, we adopt the convention that the notation \( K \) and \( D \) may also represent the set of indices rather than explicit data points, such that \( i \in K \) if and only if \( (x_i, y_i) \in K \), and the full dataset is indexed as \( D = \{1, \dots, N\} \). 

Our method consists of four key steps. First, we perform sampling steps using the Reservoir Sampling algorithm, which dynamically maintains a buffer of $K$ neighbors. Second, we estimate the $\mathcal{I}$ within the buffer using a traditional KNN-based approach. Third, we compute the value of each data point in the buffer based on the $\mathcal{I}$ obtained in the previous step. Finally, we derive the IM-KNN algorithm by incorporating Shapley-based valuation and weighted aggregation.

For step one, we have Reservoir Sampling, an efficient algorithm for randomly selecting $K$ items from a large dataset (or a data stream) of unknown size $N$, where $N \geq K$. The key advantage is that it processes the data in a single pass and maintains a fixed-size buffer (reservoir) of samples, ensuring each item has an equal probability of being selected, as shown in the Algorithm~\ref{algorithm Reservoir Sampling}.

\begin{algorithm}[t]
\caption{Reservoir Sampling}\label{algorithm Reservoir Sampling}
\begin{algorithmic}
\Require Memory buffer $K$, number of seen examples $N$, example $x$, label $y$.
\Function{Reservoir Sampling}{$K, N, D$}
    \If{$K > N$} 
        \State $K[N] \gets (x, y)$
    \Else
        \State $j \gets \text{Random Integer}(0, N]$
        \If{$j < |K|$}
            \State $K[j] \gets (x, y)$
        \EndIf
    \EndIf
    \State \textbf{Return} $K$
\EndFunction
\end{algorithmic}
\end{algorithm}
For step two, we estimate $\mathcal{I}$ between Reservoir samples. We aim to estimate the $\mathcal{I}$ between $X=\{x_i\}^K_{i=1}$ and $Y=\{y_i\}^K_{i=1}$, defined as:
\begin{equation}
\mathcal{I}(X; Y) = H(X) + H(Y) - H(X, Y),
\end{equation}
where $H(X)$ and $H(Y)$ are the marginal entropies of $X$ and $Y$. $H(X, Y)$ is the joint entropy of $X$ and $Y$. Since $ p_{X,Y}(x, y) $ is unknown, we approximate $\mathcal{I}$ using KNN. Here, distances between points are measured using the maximum norm (Chebyshev norm):
\begin{equation}
||z_i - z_u||_\infty = \max \left( ||x_i - x_u||, ||y_i - y_u|| \right).
\end{equation}
The k-th nearest neighbor of $z_i$ is denoted as:
\begin{equation}
z_i^k = \big(x^k_i, y^k_i \big).
\end{equation}
The distances are defined as:
\begin{align}
e_i &= 2||z_i - z^k_i||_\infty. \\
e^x_i &= 2||x_i - x^k_i||, \quad e^y_i = 2||y_i - y^k_i||.
\end{align}
Thus, the maximum norm distance is:
\begin{equation}
e_i = \max \big(e^x_i, e^y_i\big).
\end{equation}
 Define neighbor counts:
\begin{itemize}
    \item $ n^x_i $: Number of points with $ e^x_i \geq ||x_i - x_j|| $.
    \item $ n^y_i $: Number of points with $ e^y_i \geq ||y_i - y_j|| $.
\end{itemize}
The $\mathcal{I}$ estimator is:
\begin{equation}
\mathcal{I}(X; Y) = \psi(N) - \frac{1}{N} \sum_{i=1}^{N} \left( \psi(n^x_i) + \psi(n^y_i) \right) +\psi(k) - \frac{1}{k}, 
\end{equation}
where $ \psi(n) $ is the digamma function, defined as:
\begin{equation}
\psi(n) = \frac{d}{dn} \ln \Gamma(n) = \frac{\Gamma'(n)}{\Gamma(n)},
\end{equation}
which approximates the logarithm of a factorial. The $\mathcal{I}$ Estimator is summarized in Algorithm~\ref{Mutual Information Estimator}.

\begin{algorithm}[t]
\caption{Mutual Information Estimator}\label{Mutual Information Estimator}
\begin{algorithmic}
\Function{$\mathcal{I}$ Estimation}{$X, Y, k$}
    \State Reservoir Sampling \big($K, N, D$\big)
    \State $N \gets |X|$
    \State Initialize $e_i = 0$, $n^x_i = 0$, $n^y_i = 0$, $k$ is number of neighbors,  for $i = 1, \dots, N$
    \For{$i = 1$ to $N$}
        \State $e^x_i \gets 2  ||x_i - x^k_i||$
        \State $e^y_i \gets 2  ||y_i - y^k_i||$
        \State $e_i \gets \text{max}(e^x_i, e^y_i)$
        \State $n^x_i \gets \text{count}(||x_i - x_j|| \leq e^x_i)$
        \State $n^y_i \gets \text{count}(||y_i - y_j|| \leq e^y_i)$
    \EndFor
    \State $\mathcal{I}(X; Y) = \psi(N) - \frac{1}{N} \sum_{i=1}^{N} \left( \psi(n^x_i) + \psi(n^y_i) \right)$
    \State $\qquad \qquad + \psi(k) - \frac{1}{k}$
    \State \textbf{Return} $\mathcal{I}(X; Y)$
\EndFunction
\end{algorithmic}
\end{algorithm}
For step three, we compute the Monte Carlo Shapley value. The Shapley value of a sample $x_i$ in dataset $D$ is:
\begin{equation}
\mathcal{X}_i = \frac{1}{|D|} \sum_{K \subseteq D \setminus \{i\}} \frac{\mathcal{I}(K \cup \{i\}) - \mathcal{I}(K)}{\binom{|D|-1}{|K|}}.\label{shapley value}
\end{equation}
Here, $\mathcal{I}(K)$ is the performance score black-box oracle that takes as input any predictor and returns a score. Basically, $\mathcal{I}(K)$ is the reward if the players in the subset $K$ work together. $\mathcal{I}(K \cup \{i\}) - \mathcal{I}(K)$ is the marginal contribution of $x_i$. Computing the Shapley value in its exact form presents a significant computational challenge due to its factorial complexity with respect to the number of data points \( n \). Instead, it can be estimated using a Monte Carlo sampling approach~\cite{ghorbani19c}.

The Monte Carlo approach randomly generates multiple permutations of the dataset. Then, traverse each permutation sequentially, computing the marginal contribution of each data point. Repeat the process across numerous sampled permutations and calculate the mean marginal contribution to estimate the Shapley value.
This method leverages the equivalence to Eq.~\ref{shapley value}:

\begin{equation}
    \mathcal{X}_i = \mathbb{E}_{\mathcal{P} \sim \Pi} \left[ \mathcal{I}(K_{\mathcal{P}}^{i} \cup \{i\}) - \mathcal{I}(K_{\mathcal{P}}^{i}) \right],
\end{equation}
where \( \mathcal{P} \) denotes the uniform distribution over all \( n! \) possible permutations of the dataset, and \( K_{\mathcal{P}}^{i} \) represents the subset of data points that precede \( i \) in a randomly sampled permutation \( \mathcal{P} \). If \( i \) is the first element in \( \mathcal{P} \), then \( K_{\mathcal{P}}^{i} = \emptyset \).

This Monte Carlo-based approach provides an unbiased estimate of the Shapley value. Empirically, the estimation converges with approximately \( \mathcal{O}(n) \) samples, and three times the dataset size (\(3n\)) is often sufficient to achieve stable convergence. The Monte Carlo Shapley Value is summarized in Algorithm~\ref{Monte Carlo Shapley Value}.

\begin{algorithm}[tbp]
\caption{Monte Carlo Shapley Value}\label{Monte Carlo Shapley Value}
\begin{algorithmic}
\Function{Monte Carlo Shapley} {$K,N, D, \epsilon$}
    \State $\mathcal{I}$ ESTIMATION \big({$X, Y, k$}\big)
    \State Initialize $\mathcal{X}_i = 0$ for $i = 1, \dots, K$, and $t=0$
    \While {Convergence criteria are not met} 
        \State $t \gets t+1$
        \State $\mathcal{P}^t \gets \text{Random Permutation of data points}$
        \State $I_0^t \gets \mathcal{I}(\emptyset,\mathcal{K})$
        \For{$j = 1$ to $K$}
            \If{$|\mathcal{I}(\{\mathcal{P}^t[1], \dots, \mathcal{P}^t[j]\}) - I_{j-1}^t| < \epsilon$}
                \State $I_j^t \gets I_{j-1}^t$
            \Else
                \State $I_j^t \gets \mathcal{I}(\{\mathcal{P}^t[1], \dots, \mathcal{P}^t[j]\},\mathcal{K})$
            \EndIf
            \State $\mathcal{X}_{\mathcal{P}^t[j]} \gets \frac{t-1}{t}\mathcal{X}_{\mathcal{P}^{t-1}[j]} + \frac{1}{t}(I_j^t - I_{j-1}^t)$
        \EndFor
    \EndWhile
    \State \textbf{Return} $\{\mathcal{X}_i\}_{i=1}^K$
\EndFunction
\end{algorithmic}
\end{algorithm}

For Step four, we calculate the Euclidean distance ($d$) between the unknown data sample label and those in the Reservoir to combine weighted KNN with Shapley-based labeling. The IM-KNN is summarized in Algorithm~\ref{IM-KNN Algorithm}. The weights are computed as follows:
\begin{equation}
W_{i} = \frac{1}{(K-1)} \left(1 - \frac{ d(U_i,D_k)}{\sum_{k}d(U_i,D_k)} \right), \label{weight}
\end{equation}
where, $W_{ik}$ is the weight of sample $i$. $d(U_i,D_k)$ is the distance between unknown sample $U_i$ and neighbor $D_k$. The output label is calculated by combining weights and Shapley value as follows:
\begin{equation}
    O~\gets~\alpha(\sum_{k=1}^K W_{i} Y_{k}) + (1-\alpha)(\sum_{k=1}^K \mathcal{X}_i Y_k).
\end{equation}

\begin{algorithm}[t]
\caption{IM-KNN Algorithm}\label{IM-KNN Algorithm}
\begin{algorithmic}
\State \textbf{Input}: $D, U_i$
    \State Monte Carlo Shapley \big($K, N, D,\epsilon$\big)
    \For{$i = 1$ to $k$}
        \State $U_i \gets$ unknown sample, $D_i \gets$ known samples
        \State Calculate Euclidean distance $d(U_i, D_i)$
        \State Calculate $W_{i} \gets (1-\frac{d(U_i,D_k)}{\sum_{k}d(U_i,D_k)})/(k-1)$
    \EndFor
    \State $O$~$\gets$~$\alpha(\sum_{k=1}^K W_{i} Y_{k}) + (1-\alpha)(\sum_{k=1}^K ,\mathcal{X}_i Y_k)$
    \State \Return $O$
\State \textbf{Output}: Label
\end{algorithmic}
\end{algorithm}
\begin{table}[tbp]
\centering
\caption{Summary of datasets and their characteristics.}
\begin{adjustbox}{width=\linewidth}\label{datasets}
\begin{tabular}{cccr}

Dataset & Number of Attributes & Number of Instances  & References \\
\hline
Monk 1    & 6 & 556 &~\cite{Lichman2013}
\\\hline
Monk 2    & 6 & 601 &~\cite{Lichman2013}
\\\hline
Monk 3    & 6 & 554 &~\cite{Lichman2013}
\\\hline
Wine      & 13 & 178 &~\cite{miscwine}
\\\hline
Iris      & 4 & 150 &~\cite{misciris}
\\\hline
Bupa      & 6 & 345 & ~\cite{miscliver}
\\\hline
SAHeart   & 10 & 462 & ~\cite{saheart}
\\\hline
Wind Speed Prediction &  9  &  6574  &~\cite{fedesoriano2022wind}
\\\hline
Heart Failure Prediction &  12  &  299  &~\cite{chicco2020machine}
\\\hline
Heart Disease Prediction  & 13  & 270&~\cite{misc}
\\\hline
Heart Attack Possibility &  13  &  303  &~\cite{bhat2020health}
\\\hline
Stroke Prediction Dataset & 11  & 5110&~\cite{strokedataset}
\\\hline
MERFISH-Moffit  SRT-RNA& 155 & 64373&~\cite{Moffitt2018}
\\\hline
STARmap-AllenVISp SRT-RNA&1020 & 1549 &~\cite{Wang2018}
\\\hline
MERFISH-Moffit  SC-RNA & 18646 & 31299&~\cite{Moffitt2018}
\\\hline
STARmap-AllenVISp SC-RNA & 34617 & 14249 &~\cite{Wang2018}

\end{tabular}
\end{adjustbox}

\end{table}
\section{Results \& Discussion}
\begin{table}[t]
\centering
\caption{Accuracy (\%) comparison among KNN variants.}
\begin{adjustbox}{width=\linewidth}\label{accuracy}
\begin{tabular}{lccccccc}

\toprule
Dataset  & KNN  & L-KNN & G-KNN & M-KNN & E-KNN & IM-KNN \\
\midrule
Monk 1    & 85.31±2.14 &83.41±1.78 &86.59±2.31&90.91±1.95&94.21±1.42&\textbf{98.59±0.87}&  \\
Monk 2    & 69.91±3.12 & 75.61±2.67&69.74±3.45&70.09±2.89&80.71±2.23&\textbf{93.13±1.15} \\
Monk 3    & 89.35±1.89 & 90.51±1.56 &90±1.73&91.09±1.64&89.89±1.82&\textbf{96.52±0.93}\\
Wine      & 83.33±2.45 & 76.6±3.21 & 83.9±2.38 & 89.4±1.87 & 92.31±1.34 & \textbf{98.68±0.76} & \\
Iris       & 95.83±1.12 & 96.01±1.08 & 95.90±1.15 & 99.29±0.45 & 99.11±0.52 & \textbf{100±0.01} & \\
Bupa      & 64.51±3.67 & 65.96±3.54 &73.20±2.98&79.10±2.34&75.43±2.78&\textbf{91.07±1.45}  \\
SAHeart   & 68.43±3.23& 79.83±2.45 &73.44±2.89&76.37±2.67&78.11±2.51& \textbf{88.15±1.78}  \\
Wind Speed &  80.94±2.67  &  81.12±2.63&89.42±1.89&90.29±1.82&88.64±1.95&\textbf{99.23±0.41}\\
Heart Failure& 58.83±4.12 & 58.88±4.09 & 62.39±3.78 & 66.61±3.34 & 68.71±3.15 & \textbf{85.29±2.12} & \\
Heart Disease   &  81.35±2.56 & 77.97±2.89&79.66±2.71&82.33±2.48&81.15±2.59&\textbf{94.57±1.23}\\
Heart Attack  &76.35±2.98&71.02±3.45&68.80±3.67&76.50±2.96&77.00±2.91&\textbf{95.03±1.18} &\\
Stroke Prediction &  76.23±2.99 & 76.88±2.92&77.40±2.87&76.23±2.99&86.13±2.01& \textbf{91.70±1.35}\\
\bottomrule
Average & 77.53±2.71& 77.81±2.66&79.20±2.52&82.35±2.23&84.28±2.02&\textbf{94.33±1.12}
\end{tabular}
\end{adjustbox}
\end{table}

\begin{table}[t]
\centering
\caption{Precision (\%) comparison among KNN variants.}
\begin{adjustbox}{width=\linewidth}\label{Precision}
\begin{tabular}{lcccccc}
\toprule
Dataset  & KNN  & L-KNN & G-KNN & M-KNN & E-KNN & IM-KNN \\
\midrule
Monk 1      &95.52±1.24&92.06±1.56&88.99±1.89&92.12±1.55&98.9±0.67&\textbf{99.92±0.07}\\
Monk 2    & 72.98±3.01& 73.25±2.98&70.42±3.34&71.78±3.12&78.12±2.56&\textbf{94.54±1.21} \\
Monk 3    & 96.98±0.98& 97.06±0.96&98.09±0.73&97.42±0.89&98.99±0.63&\textbf{99.80±0.13} \\
Wine      & 83.98±2.34& 81.25±2.67&83.99±2.33&90.48±1.67&88.57±1.89&\textbf{95.00±1.12} \\
Iris      & 97.38±0.89&\textbf{100±0.01}&99.78±0.11&\textbf{100±0.02}&\textbf{100±0.00}&\textbf{100±0.01} \\
Bupa      & 40.13±4.47&46.51±4.12&61.17±3.56&63.48±3.41&78±2.45&\textbf{93.41±1.34} \\
SAHeart   & 73.49±2.89&79.52±2.34&74.29±2.82&74.87±2.76&76.67±2.61&\textbf{88.61±1.67} \\
Wind Speed  &77.42±2.78&81.78±2.41&79.09±2.63&78.48±2.71&80.96±2.47&\textbf{87.52±1.89}\\
Heart Failure  &39.13±4.34&43.21±4.12&47.01±3.89&50±3.67&54±3.43&\textbf{83.06±2.23}\\
Heart Disease   &82.98±2.43&77.55±2.87&80±2.56&74.52±3.12&85.13±2.12&\textbf{90.41±1.56}\\
Heart Attack  &81±2.51&77.40±2.78&76.50±2.89&73.89±3.07&79.24±2.63&\textbf{96.10±0.98} \\
Stroke Prediction &62±3.45&63.21±3.37&62.78±3.41&66.63±3.21&71.51±2.93&\textbf{79.43±2.34}\\
\bottomrule
Average &75.24±2.61&76.06±2.56&76.84±2.54&77.80±2.47&82.51±2.16&\textbf{92.32±1.19}\\
\end{tabular}
\end{adjustbox}
\end{table}

\begin{table}[t]
\centering
\caption{Recall (\%) comparison among KNN variants.}
\begin{adjustbox}{width=\linewidth}\label{Recall}
\begin{tabular}{lcccccc}
\toprule
Dataset  & KNN  & L-KNN & G-KNN & M-KNN & E-KNN & IM-KNN \\
\midrule
Monk 1    & 82.71±2.43 &84.29±2.31 &85.97±2.18&84.33±2.29&86.56±2.07&\textbf{91.23±1.54}  \\
Monk 2    & 70.45±3.34 & 73.22±3.12&70.34±3.36&72.19±3.18&80.28±2.43&\textbf{93.93±1.23} \\
Monk 3    & 90.02±1.78 & 90.96±1.67 &90.31±1.74&91.67±1.58&89.45±1.82&\textbf{96.49±0.89}\\
Wine      & 84.12±2.31 & 77.89±2.89 &84.68±2.28&90.05±1.78&92.97±1.42&\textbf{99.34±0.45} \\
Iris       & 95.62±1.15 & 96.18±1.07 &95.75±1.12&99.41±0.47&99.32±0.51&\textbf{99.80±0.06} \\
Bupa      & 65.73±3.51 & 65.08±3.57 &74.51±2.89&79.87±2.34&80.26±2.31&\textbf{86.02±1.89} \\
SAHeart   & 69.28±3.12&78.59±2.56&72.95±2.98&75.89±2.73&78.76±2.51&\textbf{93.15±1.34} \\
Wind Speed  &81.67±2.51&92.74±1.43&88.12±1.89&90.01±1.78&92.74±1.43&\textbf{97.00±0.78}\\
Heart Failure  &38±4.47&60.40±3.56&65.09±3.31&25±4.47&57.15±3.67&\textbf{83.29±2.23} \\
Heart Disease &70.41±3.34&69.60±3.41&69.86±3.39&78.57±2.56&70.23±3.36&\textbf{84.09±2.12}\\
Heart Attack &80±2.56&73.12±3.12&71.11±3.31&83.36±2.34&82.61±2.41&\textbf{93.51±1.34}\\
Stroke Prediction &76.08±2.89&76.74±2.84&77.17±2.81&76.08±2.89&85.88±2.12&\textbf{89.93±1.73}\\
\bottomrule
Average &75.34±2.70&78.23±2.55&77.98±2.58&78.87±2.51&83.01±2.18&\textbf{92.32±1.24}
\end{tabular}
\end{adjustbox}
\end{table}
\textbf{Datasets:} This study comprehensively evaluates KNN variants, including IM-KNN, across 12 benchmark datasets and four large-scale datasets with varying sparsity levels (60.6\%–85.6\%). The characteristics of datasets are shown in Table~\ref{datasets}.

\textbf{Training:} The experiments assess performance metrics (accuracy, precision, recall) for different $K$ values (3, 5, 7, 9, 11, 13) and validate robustness against challenges such as imbalanced data, noise, outliers, and high dimensionality. Internal validation clustering criteria—monotonicity, noise, density, subclusters, and skew distributions—are used to evaluate performance~\cite{Halkidi}. The \( R \) index is employed to measure the proximity of results to the true Number of Clusters (NC), with optimal performance indicated by minimal \( R \) values at true NC~\cite{Halkidi}. All experiments are performed over 1000 independent runs, and the average results of these experiments are reported along with the standard deviation as a confidence boundary level.

\textbf{Metric:} To validate the internal clustering of our method, we focus on five essential criteria: monotonicity, noise, density, subclusters, and skew distributions. We measure these criteria by the validity index $R$:
\begin{align}
    R = T(n) + F(n), \label{sdbw}
\end{align}
where $R$ is the validity index, and $n$  is the number of clusters. $T(n)$ is intra-cluster variance and $F(n)$ is the inter-cluster density. The $T(n)$ is  defined as:
\begin{align}
T(n) &= \frac{1}{n} \sum_{i=1}^{n} \frac{||\sigma (C_i)||}{|| \sigma (D)||}. \label{Tn}
\end{align}
Here $\sigma(C_i)$ is the variance of cluster $C_i$, and $\sigma(D)$ is the variance of a data set. The $F(n)$ defined as:

\begin{equation}
    \begin{split}
    &F(n) = \frac{1}{n(n-1)}\\ &\sum_{i}^{n}
     \sum_{j, j\neq i}^{n} \frac{\sum_{x \in C_i \cup C_j} f(x, u_{ij})}{\max\{\sum_{x \in C_i} f(x, c_i), \sum_{x \in C_j} f(x, c_j)\}}.
    \end{split}
\end{equation}
Here, $C_i$ denotes the $i-th$ cluster, and $c_i$ is the center of cluster $C_i$. $u_{ij}$ is the middle point of the line segment defined by the
clusters’ centers $c_i$, $c_j$. The neighborhood of a data point \( u \) is defined to be a hypersphere with center \( u \) and radius of the average standard deviation of the clusters. More specifically, the function \( f(x,u) \) is defined as:
\begin{align}
f(x,u) = 
\begin{cases}
    0, & \text{if } L(x,u) >  \text{stdev} \\
    1, & \text{otherwise}
\end{cases} \label{fxu}
\end{align}
If a point's distance from $u$ is less than the average standard deviation of clusters, it is evident that it belongs to the neighborhood of $u$.

\begin{figure}[t]
    \centering
    \includegraphics[width=\linewidth]{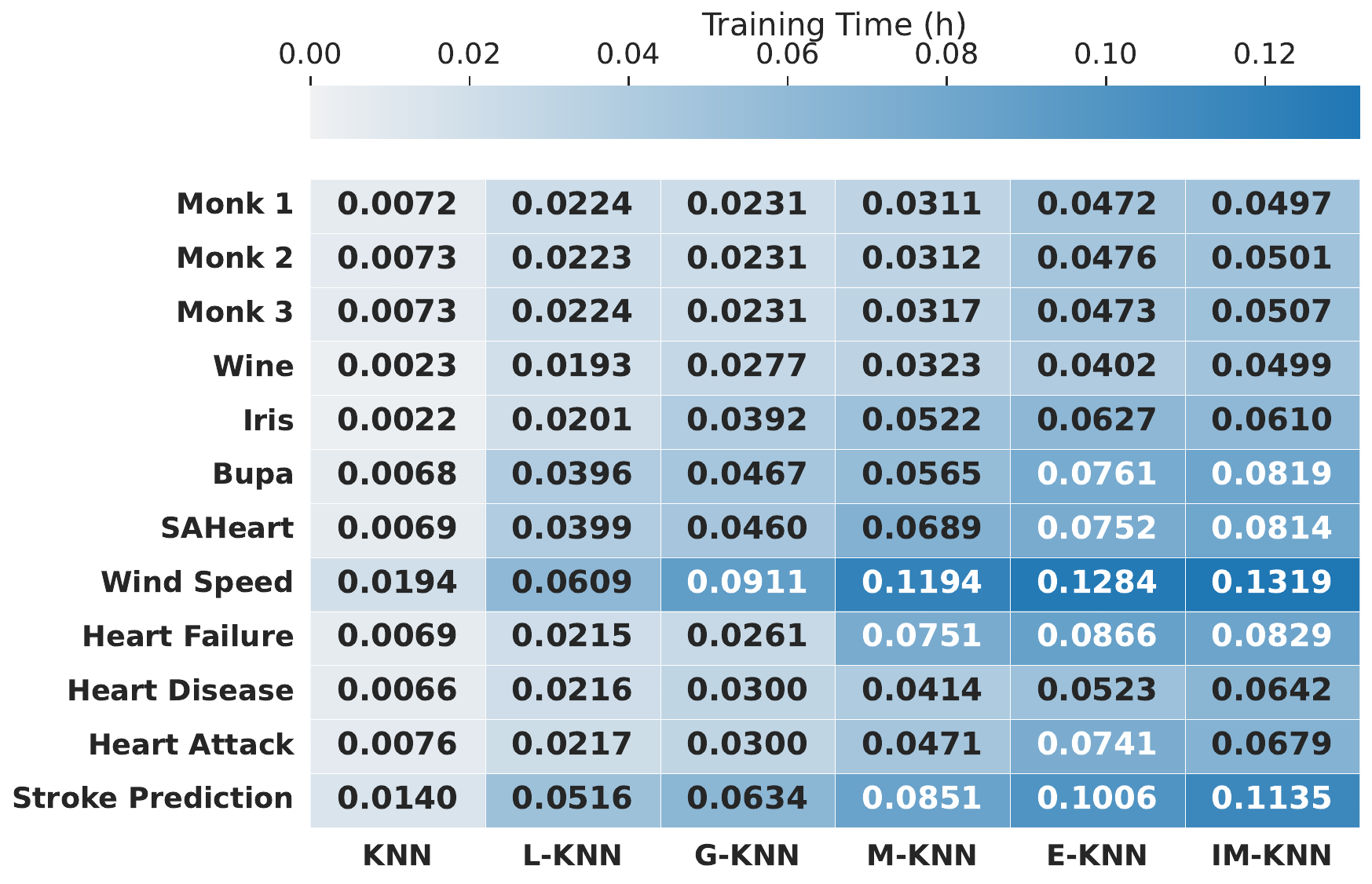}
    \caption{Training time of all methods for small-scale datasets.}
    \label{fig:aLL_METHODS_Training}
\end{figure}
\begin{figure}[t]
    \centering
    \includegraphics[width=\linewidth]{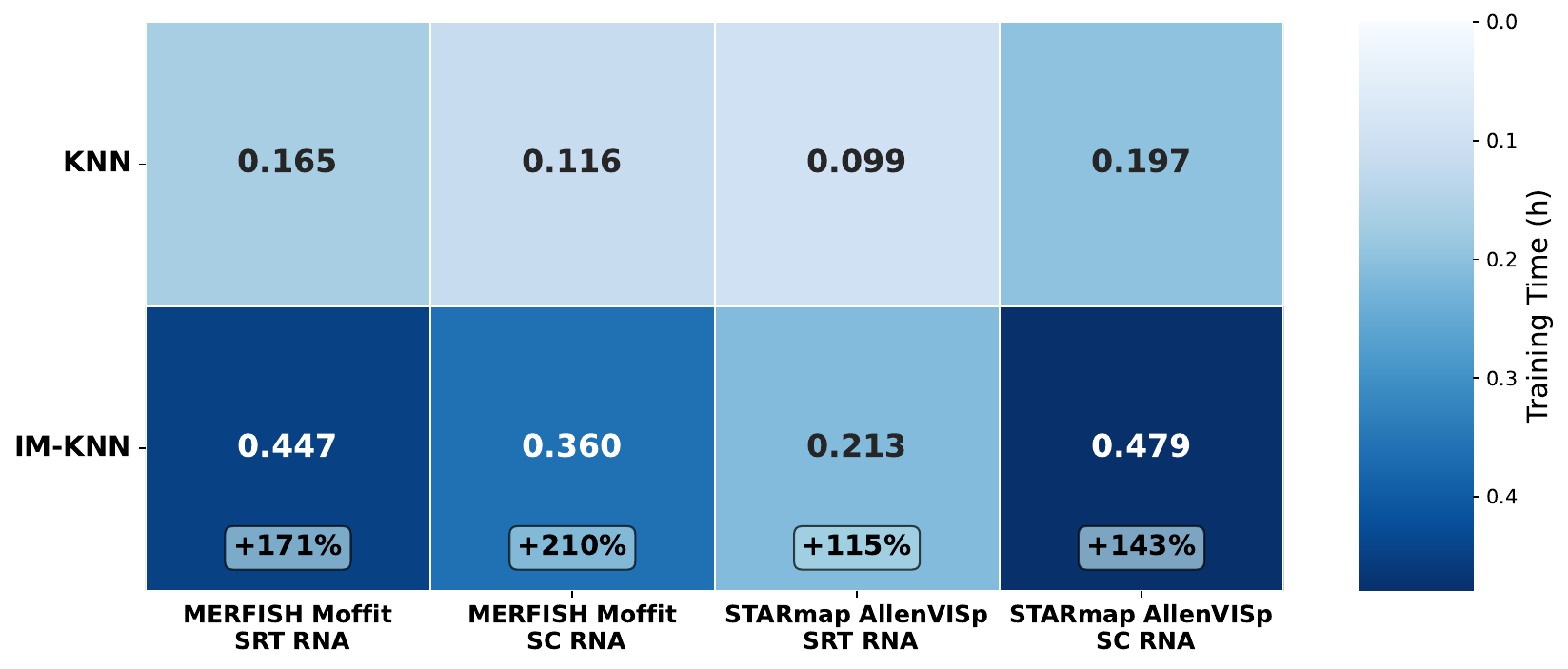}
    \caption{Training time of KNN and IM-KNN for large-scale datasets.}
    \label{fig:training time 2 methods}
\end{figure}
\begin{figure}[t]
    \centering
    \includegraphics[width=\linewidth]{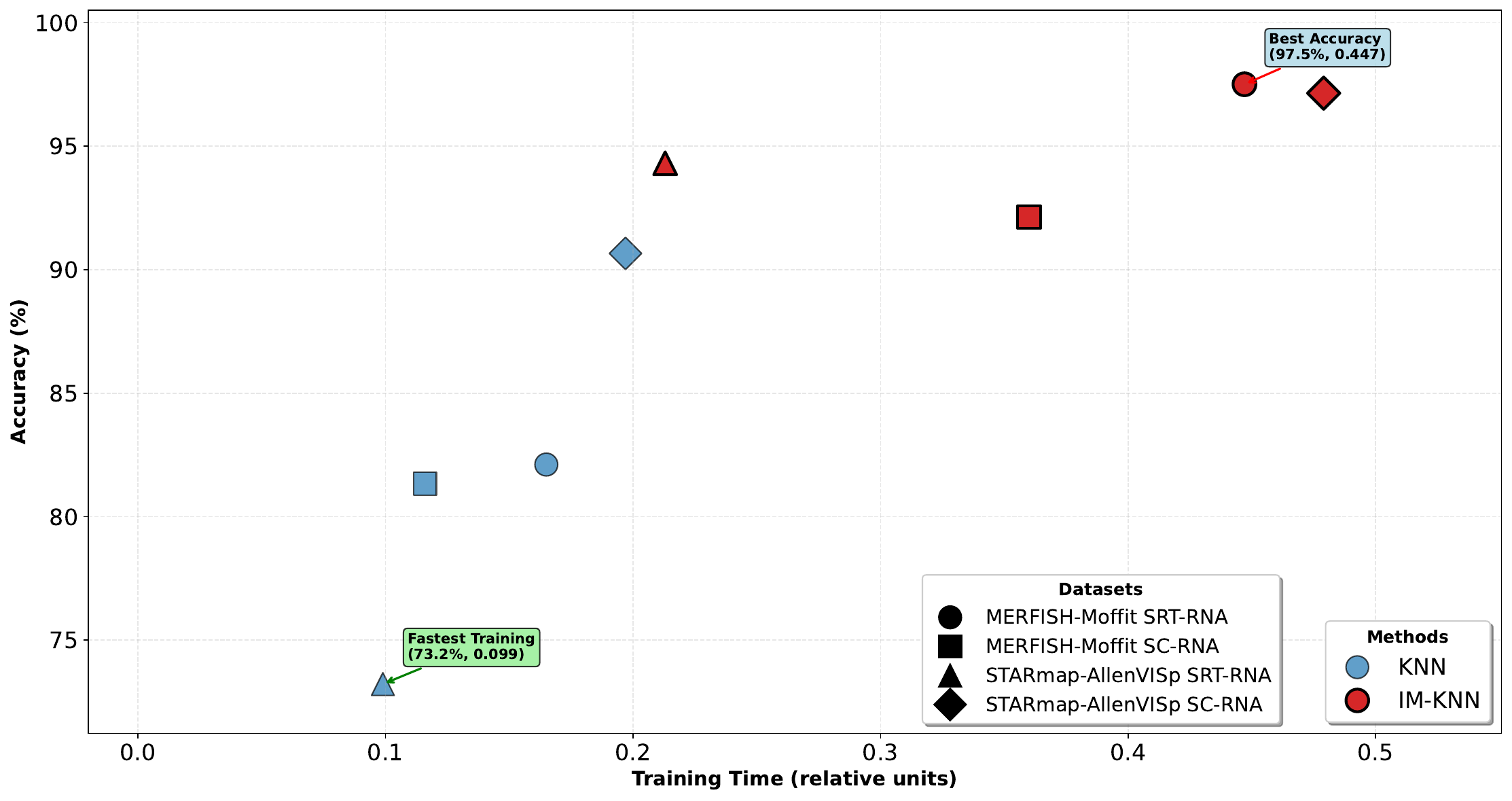}
    \caption{Training time vs Accuracy of KNN and IM-KNN for large-scale datasets.}
    \label{fig:timevsaccuracy1}
\end{figure}
\begin{table*}[t]
\centering
\caption{Comparison among KNN and IM-KNN.}
\begin{adjustbox}{width=\linewidth}\label{Sdbwknn}
\begin{tabular}{lccccccccccc}
\toprule
Dataset  & MERFISH-Moffit  SRT-RNA  & MERFISH-Moffit  SC-RNA & STARmap-AllenVISp SRT-RNA &  STARmap-AllenVISp SC-RNA  \\
\midrule
Methods & KNN\quad\qquad  IM-KNN & KNN\quad\qquad IM-KNN & KNN\quad\qquad  IM-KNN & KNN\quad\qquad  IM-KNN \\
\midrule
Accuracy         &82.11±2.41\%,  \quad\textbf{97.51±0.89\%} & 81.34±2.45\%,  \quad\textbf{92.13±1.56\%} & 73.22±3.12\%,  \quad\textbf{94.31±1.23\%} & 90.66±1.67\%,  \quad\textbf{97.15±0.91\%} &   \\
Precision        & 81.44±2.43\%, \quad\textbf{97.51±0.89\%} & 80.94±2.47\%,  \quad\textbf{90.23±1.71\%} & 77.08±2.78\%,  \quad\textbf{98.21±0.67\%} & 83.29±2.23\%,  \quad\textbf{96.51±0.98\%} & \\
Recall           & 79.39±2.61\%, \quad\textbf{98.62±0.71\%} & 83.10±2.29\%,  \quad\textbf{94.36±1.34\%} & 74.04±2.98\%,  \quad\textbf{98.00±0.78\%} & 87.80±1.89\%,  \quad\textbf{99.07±0.56\%} & \\
$R$ (The Impact of Monotonicity) & $\times$ (0.0141±0.01), \quad\textbf{$\checkmark$ (0.0022±0.01)}&$\times$ (0.0293±0.02), \quad\textbf{$\checkmark$ (0.0006±0.01)}&$\times$ (0.1004±0.05), \quad\textbf{$\checkmark$ (0.0107±0.01)}&$\times$ (0.1631±0.08), \quad\textbf{$\checkmark$ (0.0097±0.01)}& \\
$R$ (The Impact of Noise) &  $\times$ (0.8039±0.12), \quad\textbf{$\checkmark$ (0.0971±0.03)}&$\times$ (0.7214±0.15), \quad\textbf{$\checkmark$ (0.0113±0.01)}&$\times$ (0.9100±0.18), \quad\textbf{$\checkmark$ (0.0493±0.02)}&$\times$ (0.9821±0.21), \quad\textbf{$\checkmark$ (0.0854±0.03)}& \\
$R$ (The Impact of Density) & $\times$ (1.8410±0.34), \quad\textbf{$\checkmark$ (0.9835±0.18)}&$\times$ (1.0971±0.23), \quad\textbf{$\checkmark$ (0.2158±0.05)}&$\times$ (1.7444±0.31), \quad\textbf{$\checkmark$ (0.1334±0.04)}&$\times$ (1.6129±0.29), \quad\textbf{$\checkmark$ (0.2055±0.05)}& \\
$R$ (The Impact of Subclusters) & $\times$ (0.3911±0.07), \quad\textbf{$\checkmark$ (0.0125±0.01)}&$\times$ (0.5322±0.11), \quad\textbf{$\checkmark$ (0.0722±0.02)}&$\times$ (0.5741±0.12), \quad\textbf{$\checkmark$ (0.3034±0.06)}&$\times$ (0.4242±0.09), \quad\textbf{$\checkmark$ (0.0837±0.02)}& \\
$R$ (The Impact of Skewed Distributions) & $\times$ (0.3909±0.08), \quad\textbf{$\checkmark$ (0.0659±0.02)}&$\times$ (0.4003±0.08), \quad\textbf{$\checkmark$ (0.0115±0.01)}&$\times$ (0.6666±0.13), \quad\textbf{$\checkmark$ (0.0204±0.01)}&$\times$ (0.3184±0.06), \quad\textbf{$\checkmark$ (0.0192±0.01)}& \\
\bottomrule
\end{tabular}
\end{adjustbox}
\end{table*}

\textbf{Results:} IM-KNN outperforms other KNN variants in 12 of 12 small-scale datasets, achieving an average accuracy of 94.33\%, significantly higher than Ensemble KNN (84.28\%) and traditional KNN (77.53\%). Statistical significance is confirmed via two-sided paired Wilcoxon signed-rank tests (P-value = 1.8e-4), which is far below the 0.01\% threshold. The results for accuracy are shown for different methods in Table~\ref{accuracy}. In precision, as shown in Table~\ref{Precision}, IM-KNN leads in 12 datasets with an average of 92.32\%, surpassing Ensemble KNN (82.51\%) and traditional KNN (75.24\%; P-value = 1.9e-4). Similarly, as shown in Table~\ref{Recall}, IM-KNN achieves the highest recall (92.32\%) in 12 datasets, outperforming Ensemble KNN (83.01\%) and traditional KNN (75.34\%; P-value = 1.3e-4). The training time is shown in Fig.~\ref{fig:aLL_METHODS_Training}.

IM-KNN consistently outperforms traditional KNN on large-scale datasets, as shown in Table~\ref{Sdbwknn}. The results below highlight improvements in accuracy, precision, and recall, along with corresponding \textit{P}-values that confirm statistical significance (all well below the 0.01 threshold). 
In the MERFISH-Moffit SRT-RNA dataset (60.6\% sparsity), IM-KNN improves accuracy, precision, and recall by 15.40\%, 16.07\%, and 19.23\%, respectively. Moreover, the P-values are 1.4e-9, 1.5e-9, and 1.8e-9, respectively, which are significantly less than the 0.01 threshold. In the MERFISH-Moffit SC-RNA dataset (85.6\% sparsity), IM-KNN achieves gains of 10.79\%, 9.29\%, and 11.26\% in accuracy, precision,
and recall; the P-values are 1.3e-8, 1.7e-8, and 1.7e-8, respectively.
For the STARmap-AllenVISP SRT-RNA dataset
(79.0\% sparsity), improvements of 21.09\%, 21.13\%, and 23.96\% are observed, and the P-values are 1.2e-11, 1.4e-11, and 1.5e-11, respectively. In the STARmap-AllenVISP SCRNA
dataset (74.7\% sparsity), IM-KNN surpasses KNN by 6.49\%, 13.22\%, and 11.27\% in accuracy, precision, and recall; the P-values are 1.7e-7, 1.5e-7, and 1.5e-7, respectively.

Experimental results on both small- and large-scale datasets demonstrate the superior predictive accuracy of IM-KNN compared to other KNN variants. As illustrated in Fig.~\ref{fig:training time 2 methods}, traditional KNN achieves an average computational speedup of approximately 160\% over IM-KNN on large-scale datasets. However, as shown in Figs.~\ref{fig:timevsaccuracy} and \ref{fig:timevsaccuracy1}, IM-KNN is the most accurate method.

According to the $R$ index, IM-KNN consistently reaches the minimum optimal value, showcasing its overall efficacy, whereas KNN demonstrates a decreasing trend in these areas. When the methods do not achieve the optimal $R$ index at the true NC for different datasets, we mark them accordingly in Table~\ref{Sdbwknn}.  Overall, IM-KNN consistently outperforms traditional KNN across all tested metrics.

\section{Conclusion}
In this work, we introduced IM-KNN, a modified KNN variant that leverages $\mathcal{I}$ and Shapley Value to improve classification performance. By capturing richer data dependencies through a reservoir buffer and quantifying each sample’s contribution via $\mathcal{I}$-based Shapley Value, IM-KNN effectively enhances decision-making in nearest neighbor classification. Extensive experiments across diverse datasets demonstrate that IM-KNN consistently outperforms traditional KNN variants, achieving superior accuracy, precision, and recall.

{\small
\bibliographystyle{IEEEbib}
\bibliography{mlsp_template_camera_ready}

\begin{thebibliography}{10}

\bibitem{9838718}
Kun Ding, Chunlei Huo, Bin Fan, and Chunhong Pan,
\newblock ``{KNN} hashing with factorized neighborhood representation,''
\newblock in {\em ICCV}, 2015, pp. 1098--1106.

\bibitem{Darasay}
Shichao Zhang, Xuelong Li, Ming Zong, Xiaofeng Zhu, and Ruili Wang,
\newblock ``Efficient {KNN} classification with different numbers of nearest neighbors,''
\newblock {\em TNNLS}, vol. 29, no. 5, pp. 1774--1785, 2018.

\bibitem{BANSAL2022100071}
Malti Bansal, Apoorva Goyal, and Apoorva Choudhary,
\newblock ``A comparative analysis of k-nearest neighbor, genetic, support vector machine, decision tree, and long short term memory algorithms in machine learning,''
\newblock {\em Decision Analytics Journal}, vol. 3, pp. 100071, 2022.

\bibitem{6618910}
Paul Wohlhart, Martin Köstinger, Michael Donoser, Peter~M. Roth, and Horst Bischof,
\newblock ``Optimizing 1-nearest prototype classifiers,''
\newblock in {\em CVPR}, 2013, pp. 460--467.

\bibitem{shapley1951notes}
L.S. Shapley and Rand Corporation,
\newblock ``Notes on the {N}-person game,''
\newblock Tech. {R}ep., 1951.

\bibitem{pan2020}
Zhiwen Pan, Yu~Wang, and Yue Pan,
\newblock ``A new locally adaptive k-nearest neighbor algorithm based on discrimination class,''
\newblock {\em Knowledge-Based Systems}, vol. 204, pp. 106185, 2020.

\bibitem{gou2019}
Jiancai et~al. Gou,
\newblock ``A generalized mean distance-based k-nearest neighbor classifier,''
\newblock {\em Expert Systems with Applications}, vol. 115, pp. 356--372, 2019.

\bibitem{dhar2020}
Joydip Dhar, Ashaya Shukla, Mukul Kumar, and Prashant Gupta,
\newblock ``A weighted mutual k-nearest neighbour for classification mining,''
\newblock 2020.

\bibitem{hassanat2014solving}
Ahmad Hassanat, Mohammad Abbadi, Ghada Altarawneh, and Ahmad Alhasanat,
\newblock ``Solving the problem of the k parameter in the {KNN} classifier using an ensemble learning approach,''
\newblock {\em International Journal of Computer Science and Information Security}, vol. 12, pp. 33--39, 08 2014.

\bibitem{ghorbani19c}
Amirata Ghorbani and James Zou,
\newblock ``Data shapley: Equitable valuation of data for machine learning,''
\newblock in {\em ICML}, 2019, vol.~97, pp. 2242--2251.

\bibitem{Lundberg2018}
Scott~M. Lundberg, Gabriel~G. Erion, and Su{-}In Lee,
\newblock ``Consistent individualized feature attribution for tree ensembles,''
\newblock {\em arXiv preprint arXiv:1802.03888}, 2018.

\bibitem{Hamers2016}
Herbert Hamers, Bart Husslage, Robbert Lindelauf, and Tim Campen,
\newblock ``A new approximation method for the shapley value applied to the wtc 9/11 terrorist attack,''
\newblock Tech. {R}ep., Technical Report, 2016.

\bibitem{Lichman2013}
M.~Lichman,
\newblock ``{UCI},'' \url{http://archive.ics.uci.edu/ml}, 2013.

\bibitem{miscwine}
Stefan Aeberhard and M.~Forina,
\newblock ``{Wine},'' {UCI}, 1991.

\bibitem{misciris}
R.~A. Fisher,
\newblock ``{Iris},'' {UCI}, 1988.

\bibitem{miscliver}
``{Liver Disorders},'' {UCI}, 1990.

\bibitem{saheart}
``South africa heart disease dataset,'' Source: \url{http://statweb.stanford.edu/~tibs/ElemStatLearn/data.html}.

\bibitem{fedesoriano2022wind}
fedesoriano,
\newblock ``Wind speed prediction dataset,'' 2022,
\newblock https://www.kaggle.com/datasets/fedesoriano/wind-speed-prediction-dataset.

\bibitem{chicco2020machine}
Davide Chicco and Giuseppe Jurman,
\newblock ``Machine learning can predict survival of patients with heart failure from serum creatinine and ejection fraction alone,''
\newblock {\em BMC Medical Informatics and Decision Making}, vol. 20, no. 1, pp. 16, 2020.

\bibitem{misc}
``{Statlog (Heart)},'' {DOI}: https://doi.org/10.24432/C57303.

\bibitem{bhat2020health}
N.~Bhat,
\newblock ``Health care: Heart attack possibility,'' 2020,
\newblock https://www.kaggle.com/nareshbhat/ health-care-data-set-on-heart-attack-possibility.

\bibitem{strokedataset}
``Stroke prediction dataset,'' https://www.kaggle.com/datasets/fedesoriano/stroke-prediction-dataset.

\bibitem{Moffitt2018}
J.R. Moffitt et~al.,
\newblock ``Molecular, spatial, and functional single-cell profiling of the hypothalamic preoptic region,''
\newblock {\em Science}, vol. 362, 2018.

\bibitem{Wang2018}
X.~Wang et~al.,
\newblock ``Three-dimensional intact-tissue sequencing of single-cell transcriptional states,''
\newblock {\em Science}, vol. 361, 2018.

\bibitem{Halkidi}
Maria Halkidi and Michalis Vazirgiannis,
\newblock ``Clustering validity assessment: Finding the optimal partitioning of a data set,''
\newblock {\em ICDM}, pp. 187--194, 2001.

\end{thebibliography}
}
\end{document}